%% file: root.tex
\documentclass[letterpaper, 10 pt, conference]{ieeeconf}   
\IEEEoverridecommandlockouts                              
\usepackage[T1]{fontenc}
\usepackage{subcaption} 
\usepackage{graphics} 
\usepackage{amsmath} 
\usepackage{amssymb}  
\usepackage{graphicx}
\usepackage{booktabs}
\usepackage{multirow} 
\usepackage[colorlinks=true, urlcolor=blue, linkcolor=blue, citecolor=green]{hyperref}
\usepackage{amsmath,amsfonts,bm}
\usepackage[export]{adjustbox}
\usepackage[dvipsnames]{xcolor}
\definecolor{myyellow}{HTML}{F8E71C}
\definecolor{myred}{HTML}{D0021B}
\definecolor{myblue}{HTML}{4A90E2}
\definecolor{mygreen}{HTML}{5AA604}

\def\vmu{{\bm{\mu}}}
\def\mSigma{{\bm{\Sigma}}}

\usepackage{lipsum}
\usepackage{todonotes}

\title{\LARGE \bf
Uncertainty Matters: Structured Probabilistic Online Mapping for Motion Prediction in Autonomous Driving
}

\author{Pritom Gogoi$^{1,2}$, Faris Janjo\v{s}$^{1}$, Bin Yang$^{2}$, Andreas Look$^{1,3}$
\thanks{$^{1}$Bosch Center for AI, Germany, $^{2}$University of Stuttgart, Germany, $^{3}$Coburg University of Applied Sciences, Germany. }
\thanks{\tiny{As part of the IPCEI ME/CT the project is supported by the Federal Ministry for Economic Affairs and Energy, by the Ministry for Economic Affairs, Labor and Tourism of Baden-Württemberg and financed by the European Union – NextGenerationEU.}}
}

\begin{document}

\maketitle
\thispagestyle{empty}
\pagestyle{empty}

\input{sections/00_abstract}

\input{sections/01_introduction}

\input{sections/02_related_work}

\input{sections/03_approach}

\input{sections/04_experiments}

\input{sections/05_conclusion}

\bibliographystyle{IEEEtran}
\bibliography{IEEEabrv,bibliography_full}

\end{document}

%% file: sections/00_abstract.tex
\begin{abstract}

Online map generation and trajectory prediction are critical components of the autonomous driving perception-prediction-planning pipeline. While modern vectorized mapping models achieve high geometric accuracy, they typically treat map estimation as a deterministic task, discarding structural uncertainty. Existing probabilistic approaches often rely on diagonal covariance matrices, which assume independence between points and fail to capture the strong spatial correlations inherent in road geometry. To address this, we propose a structured probabilistic formulation for online map generation. Our method explicitly models intra-element dependencies by predicting a dense covariance matrix, parameterized via a Low-Rank plus Diagonal (LRPD) covariance decomposition. This formulation represents uncertainty as a combination of a low-rank component, which captures global spatial structure, and a diagonal component representing independent local noise, thereby capturing geometric correlations without the prohibitive computational cost of full covariance matrices. Evaluations on the nuScenes dataset demonstrate that our uncertainty-aware framework yields consistent improvements in online map generation quality compared to deterministic baselines. Furthermore, our approach establishes new state-of-the-art performance for map-based motion prediction, highlighting the critical role of uncertainty in planning tasks. Code is published under \url{link-available-soon}.

\end{abstract}

%% file: sections/01_introduction.tex
\section{INTRODUCTION}
\label{sec:intro}

\begin{figure}[t!]
    \centering
    \includegraphics[width=0.95\linewidth]{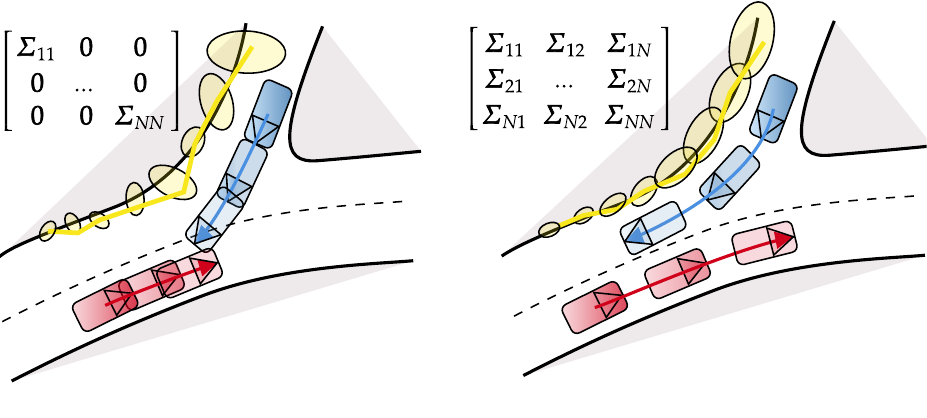}\vspace{-5pt}
    \caption{Illustration of \textcolor{myyellow}{different uncertainty representations} in probabilistic mapping for a scene with an \textcolor{myred}{ego vehicle} and a \textcolor{myblue}{predicted vehicle}. \textbf{(Left)} Treating each point within a map element (i.e. a polyline) in an independent manner can result in inconsistent covariance estimates across consecutive points. This stems from the inability to model naturally-present correlations between locations of points, such as a usually higher uncertainty for farther points. A limitation of the fully-diagonal covariance matrix representation present in the literature, this can result in erroneous predictions that in turn impact downstream planning. \textbf{(Right)} A representation that captures correlations between individual points models the underlying uncertainty more accurately. This is enabled by a covariance matrix with off-diagonal elements, which equips the probabilistic model with ability to reason about relationships between points.}
    \label{fig:dependence_illustration}
    \vspace{-5pt}
\end{figure}

Safe, comfortable, and efficient autonomous driving requires a comprehensive and reliable understanding of both the static environment and the future motion of surrounding agents. Traditionally, this understanding has relied on pre-built \textit{High-Definition} (HD) maps, which provide detailed geometric priors. However, the field is increasingly shifting toward \textit{Online Map Generation} (OMG) to handle dynamic environmental changes, reduce maintenance costs, and scale to unmapped regions~\cite{liao2023maptr}. Unlike offline maps, OMG relies on real-time sensor data that is inherently imperfect. Errors caused by sensor noise, occlusion, limited perception range, and calibration issues can propagate downstream, severely degrading the performance of safety-critical trajectory prediction and planning modules.

While modern vectorized mapping models such as MapTR~\cite{liao2023maptr} achieve impressive geometric accuracy, they typically formulate map estimation as a deterministic regression task. This approach discards valuable information regarding prediction confidence, forcing downstream planners to treat a lane boundary predicted from sparse, distant LiDAR points with the same trust as one observed at close range. 

Recognizing this gap, recent works have begun to explore probabilistic map generation. In particular,~\cite{GuSongEtAl2024a, GuSongEtAl2024_BEV, zhang2025delving} demonstrated that equipping map generators with uncertainty estimates can significantly improve the accuracy of trajectory prediction. However, these existing approaches typically rely on simplified uncertainty representations, such as diagonal covariance matrices. This assumption of independence treats a map element as a collection of unrelated points, failing to capture the spatial correlations inherent in road geometry. For instance, the curvature of a lane segment implies that positional errors are highly correlated across its points; if one point is shifted due to calibration noise, its neighbors are likely shifted similarly. This is illustrated in Fig.~\ref{fig:dependence_illustration}.

\begin{figure}[h!]
\begin{subfigure}[t]{0.5\textwidth}
\centering
\includegraphics[width=0.8\linewidth,fbox]{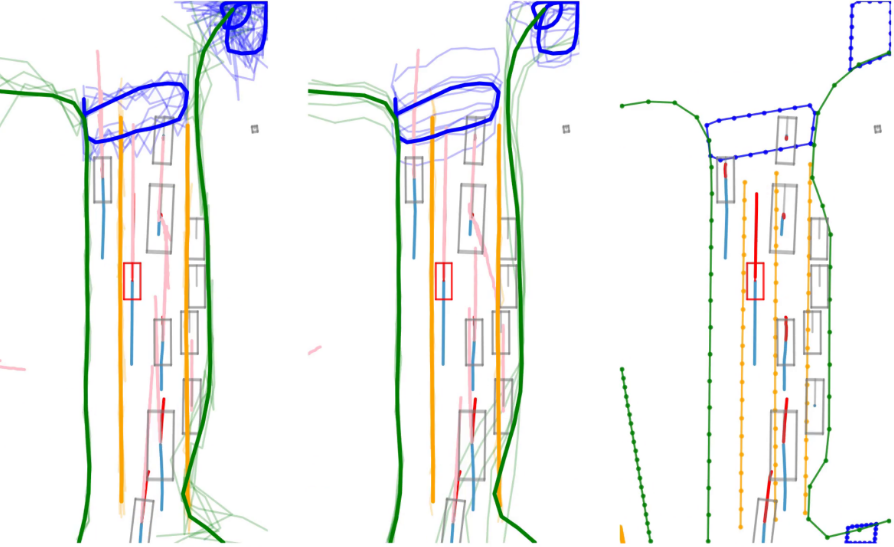}
\end{subfigure}


\begin{subfigure}[t]{0.5\textwidth}
\centering  
\includegraphics[width=0.8\linewidth, trim={0 3cm 0 3cm},clip,fbox]{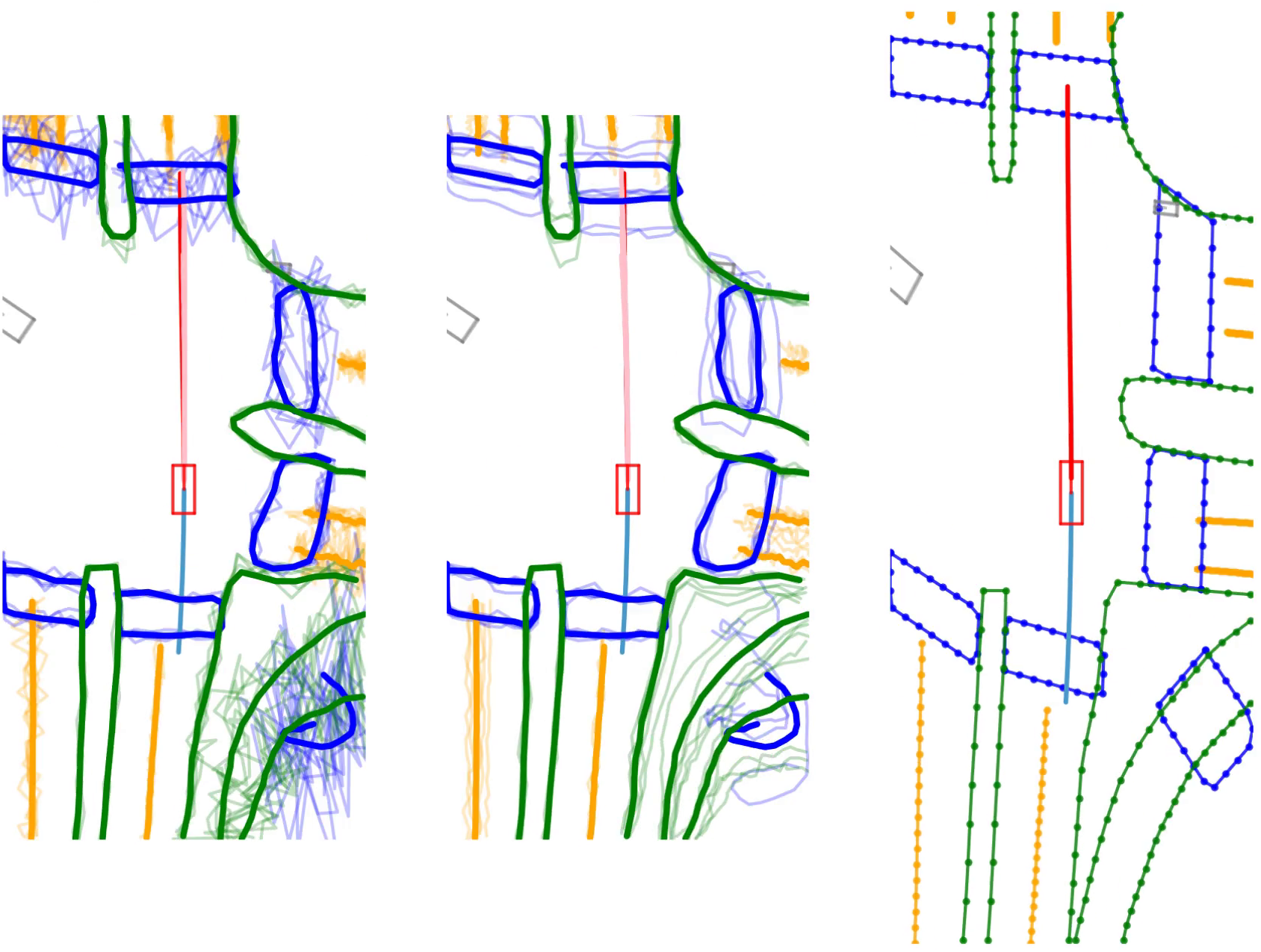}
\label{fig:subfig_b}
\end{subfigure} 
    \caption{Qualitative comparison of OMG predictions on several nuScenes~\cite{nuscenes} scenarios. We display samples drawn from the predicted distribution as thin transparent lines, while the thick solid line represents the predictive mean.
    \textbf{(Left)} Baseline models using diagonal covariance (independent uncertainty) fail to capture dependencies between points, resulting in jagged, spatially incoherent samples.
    \textbf{(Middle)} Our proposed LRPD model explicitly accounts for spatial correlations, producing smooth and geometrically consistent samples.
    \textbf{(Right)} The Ground Truth HD map for reference.}
    \label{fig:qualitative_map_uncertainty}
\vspace{-10pt}
\end{figure}

Neglecting these correlations results in "jagged," spatially incoherent uncertainty estimates that fail to represent the true geometric ambiguity of the scene. We visualize this issue in Fig.~\ref{fig:qualitative_map_uncertainty}, where samples drawn from a diagonal covariance model (left) exhibit unrealistic high-frequency oscillations. In contrast, samples drawn from our method (middle) are smooth and geometrically plausible.

To address these limitations, we present a structured probabilistic framework for online map generation. Unlike prior methods that model point-wise variance in isolation, we propose a formulation that explicitly models the dependencies within map elements using a \textit{Low-Rank plus Diagonal} (LRPD) covariance decomposition. This allows the system to reason about the uncertainty.
Our key contributions are summarized as follows:
\begin{enumerate}
    \item We utilize a \textbf{fully-probabilistic formulation} of the underlying map uncertainty in OMG, without unrealistic and limiting assumptions present in existing work. This representation correctly captures spatial dependencies and geometric correlations within map elements.
    \item We propose a structured and lean interface utilizing a \textbf{low-rank plus diagonal covariance parameterization} that enables training probabilistic representations in a stable and efficient manner, without the computational cost of full covariance matrices.
    \item We benchmark our method on the nuScenes dataset, establishing \textbf{new state-of-the-art performance} for map-based motion prediction. Our experiments demonstrate that structured uncertainty modeling yields consistent gains over both deterministic baselines and methods relying on independent uncertainty assumptions.
\end{enumerate}

%% file: sections/02_related_work.tex
\section{Related Work}
\label{sec:related_work}

In this section we contextualize our approach within the shift from static to online mapping and the growing need for probabilistic reasoning in autonomous driving. 

\subsection{Online Map Generation (OMG)}
The reliance on pre-annotated HD maps restricts the scalability of autonomous driving systems due to high maintenance costs and the inability to handle dynamic map changes~\cite{wong2020mapping}. 
Consequently, online map generation systems that infer the static road network in real-time from onboard sensors, have become prominent~\cite{li2021hdmapnet, dong2024superfusion, liu_VectorMapNet, shin2025instagram, liao2023maptr, wang2023exploring, xu2024insmapper, zhang2024enhancing}. 
Early approaches primarily focused on dense, rasterized Bird's-Eye-View (BEV) semantic segmentation, lifting camera features into 3D space to predict drivable areas and lane markings~\cite{philion2020lift, li2021hdmapnet, liu2023bevfusion, ma2024vision, can2022understanding, li2024bevformer, li2023delving}.

However, downstream planning modules typically require structured geometric representations rather than dense pixel grids. To address this, vectorization methods such as MapTR~\cite{liao2023maptr},~MapTRv2~\cite{MapTRv2}, or StreamPETR~\cite{wang2023exploring} have recently been proposed. These models utilize Transformer-based architectures to directly regress vectorized map elements (polylines) from BEV features, achieving impressive geometric precision. Despite their accuracy, these state-of-the-art methods typically formulate map estimation as a deterministic regression task, discarding critical information regarding the model's confidence in its own predictions.

\subsection{Uncertainty in Mapping}
Accurate uncertainty estimation is vital for safety-critical applications. 
While probabilistic object detection is well-studied~\cite{kendall2017uncertainties}, uncertainty in vectorized map construction remains underexplored. Recent works have begun to address this gap. {Gu et al.}~\cite{GuSongEtAl2024a} proposed a probabilistic formulation for map generation, demonstrating that modeling uncertainty can improve downstream prediction performance. However, their approach is limited to independent Laplace distributions. This assumption treats map points as independent entities, ignoring the strong spatial correlations inherent in continuous road structures like lane dividers.

More recently, {Zhang et al.}~\cite{zhang2025delving} improved upon this by introducing a formulation that captures the correlation between the $x$ and $y$ coordinates of a single point. While this allows for block-diagonal covariance (capturing point-wise ellipsoidal uncertainty), it still assumes that the error at one point is independent of its neighbors. In contrast, our work explicitly models spatial dependencies {between} distinct points within a map element using a low-rank plus diagonal decomposition. This enables the system to reason about the shape and structural ambiguity of the road geometry.

\subsection{Map-Aware Trajectory Prediction}
Trajectory prediction is a core component of the autonomous driving stack, with modern approaches heavily relying on map context to forecast agent motion. To address the inherent stochasticity of human driving behavior, recent pipelines have moved beyond deterministic regression to explicitly model the multi-modal distribution of future trajectories. Generative approaches such as CVAE-based frameworks~\cite{salzmann2020trajectron++, janjovs2023conditional}, diffusion models~\cite{jiang2023motiondiffuser}, or discrete autoregressive architectures~\cite{seff2023motionlm}, have shown great success in capturing diverse agent intentions, i.e. the probability of yielding versus proceeding or changing lanes versus maintaining course. These methods typically output non-deterministic forecasts either as samples from latent distributions~\cite{janjovs2023conditional} or by directly constructing \textit{Gaussian Mixture Models}  (GMMs)~\cite{look2023_cheap, liu2024_laformer} or occupancy heatmaps~\cite{gilles2022gohome,gilles2022thomas}, effectively handling the aleatoric uncertainty of motion.

However, despite their sophisticated handling of agent dynamics, these methods typically assume perfect, ground-truth map inputs or rely on deterministic predictions from upstream mapping modules~\cite{hu2023uniad, Weng_2024_CVPR}. This creates a critical disconnect: while the agents are modeled probabilistically, the map is treated as a fixed, certain reality. This assumption can lead to catastrophic errors when the map perception is noisy or occluded, as illustrated in Fig.~\ref{fig:dependence_illustration}. For instance, a diffusion-based predictor might correctly model a vehicle's intent to follow a lane, but if the upstream mapper erroneously predicts a sharp turn due to sensor noise, the predicted trajectory will be confident yet incorrect.

%% file: sections/03_approach.tex
\section{Method}
\label{sec:approach}

Our framework addresses the problem of joint online map generation and trajectory prediction under uncertainty. We propose a two-stage pipeline: (1) a probabilistic mapping module that estimates vectorized map elements with structured geometric uncertainty, and (2) an uncertainty-aware trajectory prediction module that uses these estimates to forecast agent motions robustly.

\subsection{Problem Formulation}
Let the local environment be represented by a set of map elements $\mathcal{M} = \{\mathbf{m}_1, \dots, \mathbf{m}_K\}$. Each element $\mathbf{m}_k$ (e.g., a lane divider or road boundary) can be described as a polyline consisting of $N$ ordered points in the bird's-eye-view (BEV) plane. The polyline $\mathbf{m}_k$ is then mathematically represented as a flattened vector $\mathbf{m}_k \in \mathbb{R}^{2N}$ containing a sequence of $N$ 2D points with coordinates $(x_1, y_1, \dots, x_N, y_N)$.

Conventionally, map generators regress the coordinates of points in $\mathbf{m}_k$ deterministically. Instead, we model the conditional distribution $\mathcal{P}(\mathbf{m}_k | {\mathbf{x}})$, where ${\mathbf{x}}$ contains contextual information such as sensor inputs (e.g., surround-view images). The distribution $\mathcal{P}(\mathbf{m}_k | {\mathbf{x}})$ is represented by a multivariate Gaussian
\begin{equation}
    \mathcal{P}(\mathbf{m}_k | \mathbf{x}) = \mathcal{N}(\mathbf{m}_k |\vmu_{\phi, k}(\mathbf{x}), \mSigma_{\phi, k}(\mathbf{x}))\ ,
\end{equation}
parameterized by neural network parameters $\phi$. Here, $\vmu_{\phi, k} \in \mathbb{R}^{2N}$ is the predicted mean position of the map element, and $\mSigma_{\phi, k} \in \mathbb{R}^{2N \times 2N}$ is the covariance matrix capturing the geometric uncertainty. 

As the prediction is generated for all $K$ map elements in a scene, the overall output of the map generator can be described as a set of individual weighted distributions
\begin{align}
\begin{split}
    \mathcal{P}(\mathcal{M} | \mathbf{x}) &= \{ \left(c_{\phi, k}(\mathbf{x}), \mathcal{P}(\mathbf{m}_k | \mathbf{x})\right) \}^K_{k=1}\ \\
     &= \{ \left(c_{\phi, k}(\mathbf{x}), \mathcal{N}(\mathbf{m}_k |\vmu_{\phi, k}(\mathbf{x}), \mSigma_{\phi, k}(\mathbf{x}))\right) \}^K_{k=1}\ ,
\end{split}
\end{align} 
where $c_{\phi, k}(\mathbf{x}) \in [0, 1]^C$ resides on the standard simplex and denotes the predicted class probability vector assigned to the $k$-th map element across $C$ possible categories.

A key challenge in the formulation above is the structure of the covariance matrix $\mSigma_{\phi, k}$ for every map element. Predicting a full, dense covariance matrix involves estimating $O(N^2)$ parameters, which is not only computationally prohibitive but also prone to severe optimization instability -- a limitation we empirically demonstrate in Sec. \ref{sec:experiments}. Consequently, prior works have resorted to simplifying assumptions that sacrifice  geometric information:
\begin{itemize}
    \item \textbf{Independent Uncertainty:} Gu et al. \cite{GuSongEtAl2024a} model the output as a Laplace distribution in which each coordinate is treated independently. This assumes $\mSigma_{\phi, k}$ is strictly diagonal, ignoring all correlations.
    \item \textbf{Point-wise Correlation:} Zhang et al. \cite{zhang2025delving} extend this by using a Gaussian distribution that captures correlations between the $x$ and $y$ coordinates of a {single} point. In this case, $\mSigma_{\phi, k}$ is block-diagonal consisting of $N$ blocks of size $2 \times 2$, but correlations between {different} points are zero.
\end{itemize}

In contrast, our work explicitly models the spatial dependencies between distinct points within a map element $\mathbf{m}_k$ by assuming a full covariance matrix $\mSigma_{\phi, k}$ that predicts all $2N\times 2N$ elements. In the following, we describe how we maintain this high-capacity representation efficient.

\subsection{Structured Geometric Uncertainty}
A core challenge in vectorized mapping is modeling the spatial dependencies between points in an efficient manner. 
To address this challenge, we propose a \textit{Low-Rank Plus Diagonal} (LRPD) parameterization. We decompose the covariance matrix $\mSigma_{\phi, k}$ into a diagonal component representing independent noise and a low-rank component capturing spatial correlations
\begin{equation}
    \mSigma_{\phi, k} = \mathbf{D}_{\phi, k} + \kappa\mathbf{L}_{\phi, k} \mathbf{L}_{\phi, k}^T \ ,
\end{equation}
where
\begin{itemize}
    \item $\mathbf{D}_{\phi, k} = \text{diag}(d_{\phi, 1, k}, \dots, d_{\phi, 2N, k})$ with $d_{\phi, n,k} > 0$ represents the independent variance of each coordinate and element,
    \item $\mathbf{L}_{\phi, k} \in \mathbb{R}^{2N \times R}$ is a factor matrix with rank $R \ll 2N$,
    \item $\kappa \geq 0$ is a scalar scaling factor controlling the influence of the low-rank component during model training.
\end{itemize}
This formulation reduces the parameter space to $O(NR)$, making it tractable for real-time applications while preserving the expressivity needed to model geometric dependencies. The rank $R$ is a hyperparameter that controls the complexity of the correlation structure; in our experiments, we find $R=24$ sufficient to capture primary modes of uncertainty such as translation and curvature ambiguity. Notably, for $2N=100$, this approach reduces the total number of parameters by a factor of four.

\subsection{Alternative Uncertainty Representations}
Prior to developing the proposed LRPD architecture, we explored more direct covariance representations. Specifically, we attempted to directly predict the full covariance matrix, including all off-diagonal elements. However, this approach proved highly unstable during training, frequently resulting in predicted matrices that were not positive semi-definite. To mitigate this instability, we experimented with several numerical strategies: applying custom weight initializations for off-diagonal elements, injecting noise, scheduling the separate prediction of diagonal and off-diagonal terms, and testing various combinations of freezing and fine-tuning the means and covariances. Ultimately, none of these alternative approaches yielded stable results, motivating the design of the proposed LRPD.

\subsection{Probabilistic Learning Objective}
During training, we incentivize the map generator to maximize the log-likelihood of observing the ground-truth map coordinates $(x^*_1, y^*_1, \dots, x^*_N, y^*_N)$ given the predicted distribution. The loss function is the \textit{Negative Log-Likelihood} (NLL)
\begin{align}
    \mathcal{L}_\text{NLL} = \sum_k \log |\mSigma_{\phi, k}| +
         \mathbf{r}_k^T \mSigma^{-1}_{\phi,k} \mathbf{r}_k \ ,
\end{align}
for each dataset sample, where $\mathbf{r}_k = \mathbf{m}^*_{k} - \vmu_{\phi,k}$ represents the spatial residual, and $\mathbf{m}^*_{k}$ is the ground-truth polyline.

Following standard set prediction architectures~\cite{MapTRv2, liao2023maptr}, we also apply a classification loss to the predicted class labels $c_{\phi, k}(\mathbf{x})$. Specifically, this is implemented as a Focal Loss to encourage $c_{\phi, k}(\mathbf{x})$ to match the ground-truth class labels. 

\subsection{Map-Uncertainty-Aware Trajectory Prediction}
The predicted map distribution is consumed by a downstream trajectory prediction module. Prediction modules usually consist of sequentially-ordered encoder and decoder networks. The encoder consumes the inferred map context $\mathcal{M}$ as well as historical motion of relevant agents to construct latent features that describe the entire driving scene. These features are passed to the decoder, tasked with generating predicted future motion for agents of interest. As our inferred map representation $\mathcal{M}$ contains uncertainty, we must provide this information to the downstream encoding process. In the following, we introduce two complementary mechanisms.

\subsubsection{Explicit Uncertainty Encoding}\label{subsubsec: explicit_enc}
To explicitly incorporate geometric uncertainty, we augment the input feature space of the trajectory prediction encoder. While standard approaches rely purely on spatial coordinates, we construct an enriched feature representation for each point in the map polyline element $\mathbf{m}_k = (x_1, y_1, \dots, x_N, y_N)$. Specifically, the input feature vector $\mathbf{e}_{n, k} \in \mathbb{R}^{4+2R}$ for the $n$-th point in $\mathbf{m}_k$ is formed by concatenating:
\begin{enumerate}
    \item the predicted mean coordinates $\mu_{x, n}, \mu_{y, n} \in \mathbb{R}$,
    \item the independent diagonal variances $\sigma^2_{xx, n}, \sigma^2_{yy, n} \in \mathbb{R}_+$ extracted from the diagonal matrix $\mathbf{D}_{\phi, k}$, and
    \item the corresponding row vectors $\mathbf{l}_{x, n}, \mathbf{l}_{y, n} \in \mathbb{R}^R$ from the low-rank factor matrix $\mathbf{L}_{\phi, k}$.
\end{enumerate}
Consequently, a single map element $\mathbf{m}_k$ is represented by an $N(4+2R)$-dimensional feature vector. By providing the encoder with the low-rank rows $\mathbf{l}_{x, n}$ and $\mathbf{l}_{y, n}$, we enable the network to reason about spatial correlations and distinguish between noisy observations and precise geometric measurements at the feature level.

\subsubsection{Confidence-Aware Feature Modulation}\label{subsubsec: film}
As an extension, we introduce a mechanism to explicitly incorporate the predicted class label $c_{\phi, k}$ associated with each map element $\mathbf{m}_k$. This enables the model to factor the relative confidence of a map element's existence into the spatial uncertainty of its constituent polyline points. Following standard practices in recent literature~\cite{GuSongEtAl2024a, zhang2025delving}, we extract the predicted confidence score for a specific, predefined road type (e.g., lane centerlines) rather than utilizing the full probability simplex. This simplifies the predicted class label vector into a single scalar value. To integrate this scalar confidence, we employ \textit{Feature-wise Linear Modulation} (FiLM)~\cite{perez2018film}, which serves as an effective, principled, and lightweight mechanism to dynamically adjust the per-point lane embeddings.

First, we pass the feature vector $\mathbf{e}_{n, k}$ through a $D_e$-dim. linear layer to obtain an embedding $\tilde{\mathbf{e}}_{n,k} = f_\theta(\mathbf{e}_{n,k})$ for each point $n$ in the $k$-th map element. Then, the class label score of each map element $c_{\phi, k}$ is assigned to each point in $\mathbf{m}_k{=}(x_1, y_1, \dots, x_N, y_N)$. FiLM computes the resulting confidence-aware lane embedding $\hat{\mathbf{e}}_{n,k}$ by applying an affine transformation
\begin{equation}
    \hat{\mathbf{e}}_{n,k} = \textrm{ReLU}(\gamma(c_{\phi, k})) \odot \tilde{\mathbf{e}}_{n,k} + \beta(c_{\phi, k})\ ,
\end{equation}
where $\gamma(\cdot)$, $\beta(\cdot): \mathbb{R} \rightarrow \mathbb{R}^{D_e}$ are learnable functions implemented as linear projections, $\textrm{ReLU}(\cdot)$ is the rectified linear unit activation function and $\odot$ denotes element-wise multiplication. The modulation parameters scale and shift each feature dimension independently. This allows the confidence value to influence the contribution of lane information in a fine-grained manner. We choose $D_e=128$ and ReLU as the activation function after determining its effectiveness empirically over different activations. We also evaluated alternative strategies for incorporating the confidence term, such as direct concatenation (without the FiLM layer), attention mechanisms~\cite{vaswani2017attention}, and sinusoidal Fourier embeddings. However, FiLM-based modulation consistently yielded superior performance.

Finally, the downstream trajectory prediction module utilizing the probabilistic map information is intended to be general-purpose. We do not impose any structural constraints on the polyline encoder, other than its capacity to process the modulated features generated by the explicit uncertainty encoding and the FiLM layer.

%% file: sections/04_experiments.tex
\section{Experiments}
\label{sec:experiments}

Our experimental evaluation aims to verify two primary hypotheses: (i) that the LRPD formulation provides more stable and physically consistent map uncertainty than diagonal baselines, and (ii) that these structured correlations directly improve the robustness of downstream trajectory prediction. We compare our approach against state-of-the-art deterministic and probabilistic mapping frameworks on large-scale real-world data.

\subsection{Dataset and Preprocessing}
We evaluate our approach on the \textit{nuScenes} dataset \cite{nuscenes}, which provides 1,000 urban driving scenes across Boston and Singapore. The dataset includes comprehensive sensor modalities and high-quality vectorized map annotations.

The standard nuScenes agent annotations are provided at 2 Hz. However, high-fidelity motion forecasting requires finer temporal resolution to capture subtle maneuvering dynamics. Following the protocol in \cite{GuSongEtAl2024a}, we utilize the \texttt{trajdata} library \cite{ivanovic2023trajdata} to upsample agent trajectories to 10 Hz via linear temporal interpolation. This ensures compatibility with standard trajectory prediction benchmarks and provides a denser supervision signal for training.

\subsection{Evaluation Metrics}
We assess performance across two distinct tasks to demonstrate the utility of structured uncertainty.

\subsubsection{Map Generation Metrics}
To measure the geometric accuracy of the predicted vectorized maps, we employ the \textit{Mean Average Precision} (mAP). We represent both the predicted and the ground-truth map element as ordered sets of sampled points. The similarity between them is measured using the bidirectional Chamfer distance.
A predicted map element is considered a true positive if its Chamfer distance to a ground-truth element of the same class is within a predefined threshold. Following \cite{liao2023maptr}, we compute the average precision by integrating the precision-recall curve for each class across three distance thresholds $\{0.5 \text{m},1.0 \text{m},1.5 \text{m}\}$. The final mAP is obtained by averaging the average precision across all map categories.

\subsubsection{Trajectory Prediction Metrics}
We evaluate forecasting performance using standard distance-based metrics for the $K=6$ most likely modes:
\begin{itemize}
\item {minADE$_6$:} The average $L_2$ distance between the ground-truth and the closest predicted trajectory over the full horizon.
\item {minFDE$_6$:} The $L_2$ distance between the ground-truth position and the endpoint of the best predicted mode.
\item \textit{Miss Rate} (MR): The percentage of cases where the minFDE exceeds 2.0 meters.
\end{itemize}

\subsection{Implementation Details}
\subsubsection{Architecture}
For the base mapping architecture, we adopt MapTR~\cite{liao2023maptr} and MapTRv2~\cite{MapTRv2}. We also evaluate an extension, MapTRv2-centerline (MapTRv2-CL), which explicitly models lane centerlines as a separate map element category. To enable probabilistic mapping, we modify the standard regression heads of these architectures to output the parameters of our proposed LRPD formulation. 

For downstream motion forecasting, we utilize HiVT~\cite{zhou2022hivt}, augmented with an uncertainty-aware encoder designed to consume our structured map covariances. We explicitly prioritize HiVT over other commonly cited modules like DenseTNT~\cite{gu2021densetnt} because HiVT provides significantly lower trajectory prediction errors, serving as a highly accurate and established baseline in recent uncertainty-aware literature.

We also acknowledge the notable recent work~\cite{sun2026diffsemanticfusion} that achieves further gains by combining a computationally cumbersome raster representation obtained from BEV features with a diffusion mechanism in the prediction decoder. 
As their work does not focus on map uncertainty and draws on additional gains by the decoder, we instead aim to isolate and accurately measure the specific improvements yielded by the probabilistic map representation. Thus, we restrict our experimental baseline to the established HiVT architecture. 

\begin{figure}[t!]
    \centering
    \includegraphics[width=0.95\linewidth]{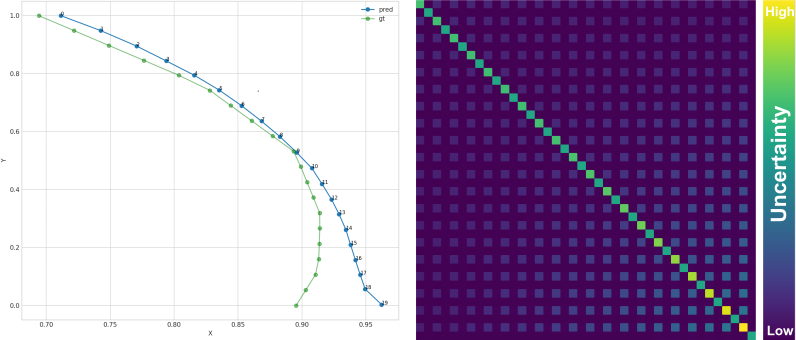}
\caption{Uncertainty calibration of our method. \textbf{(Left)} Visualization of \textcolor{blue}{predicted} versus \textcolor{mygreen}{ground-truth} map element. The prediction aligns closely with the ground truth initially, but the positional error gradually increases toward the end. \textbf{(Right)} Predicted covariance demonstrating that our model's uncertainty  correlates with the error. Uncertainty grows correspondingly as the prediction diverges. Specifically, the model correctly outputs low uncertainty for the $y$-coordinate (where the error remains small) and appropriately assigns high uncertainty to the $x$-coordinate as its error increases.}
    \label{fig:cov_illustration}
\end{figure}

\subsubsection{Training Procedure}
Our model is trained in a two-stage fashion to ensure optimization stability. We use the AdamW optimizer \cite{LoshchilovH19} with a learning rate of $6 \times 10^{-4}$ and cosine annealing schedule.

\textit{Map Generation}:
The mapping module is trained using the NLL loss. A critical challenge in learning full covariance structures is early-stage instability. To mitigate this, we introduce a curriculum scheduling parameter that weights the off-diagonal terms of the covariance.

\begin{itemize}
\item {Warmup Phase:} Initially, we set $\kappa=0$ so that the model predicts only the diagonal covariance elements $\mathbf{D}_{\phi, k}$. The learning problem reduces to predicting independent Gaussian uncertainty, which is easier to optimize.
\item {Structured Learning Phase:} After a fixed number of epochs, we gradually increase $\kappa$, allowing the model to learn spatial correlations.
\end{itemize}

\textit{Trajectory Prediction}:
The motion forecasting module is trained on the frozen outputs of the mapping stage. By keeping the map generator fixed, we ensure that the trajectory prediction head learns to interpret the structured uncertainty as a reliable signal of geometric ambiguity, rather than the two tasks competing for feature representations. This strategy is consistent with established protocols \cite{GuSongEtAl2024a}.

\subsection{Results}
\label{sec:results}

In this section, we present a quantitative evaluation of our structured uncertainty framework, focusing on map estimation quality and downstream prediction performance.

\subsubsection{Online Map Generation Performance}
Table \ref{tab:map_results} summarizes the map generation performance across different base architectures. To ensure a rigorous comparison, we reproduce results from existing literature \cite{GuSongEtAl2024a} by re-running the provided codebases within our environment. We observe that extending the training schedule from the standard 24 epochs to 45 epochs yields consistent improvements in mAP for all models. When re-running the baseline code for 24 epochs, we achieve results identical to those reported in the literature; however, for completeness, we report both the original values and our reproduced long-schedule results.

The evaluation across all three online map generation architectures reveals that equipping deterministic models with probabilistic outputs is inherently beneficial for geometric accuracy. Furthermore, incorporating structured uncertainty via our LRPD formulation provides additional gains, achieving the highest mAP across all tested configurations. These results suggest that modeling spatial correlations not only improves the representation of uncertainty but also provides a more robust supervision signal for the geometric mean itself.

Additionally, we qualitatively visualize a predicted and ground-truth map element alongside the predicted covariance in Fig.~\ref{fig:cov_illustration}. We observe that the spatial error strongly correlates with predicted uncertainty, exhibiting a highly non-diagonal covariance structure.

\begin{table}[t]
\centering
\caption{Online map generation performance for different backbones types and uncertainty estimation methods.}
\label{tab:map_results}
\begin{tabular}{lc}
\toprule
\textbf{Map Model}  & \textbf{mAP} $\uparrow$  \\ 
\midrule
\textit{Backbone: MapTR \cite{liao2023maptr}}   \\
+ Deterministic & 0.4488\\
+ Uncertainty: iid \cite{GuSongEtAl2024a} (reported) & 0.4525 \\
+ Uncertainty: iid \cite{GuSongEtAl2024a} (reproduced) & 0.5158\\
{+ Uncertainty: LRPD (ours)}  & \textbf{0.5589} \\ 
\midrule
\textit{Backbone: MapTRv2 \cite{MapTRv2}}  \\
+ Deterministic & 0.5540  \\
+ Uncertainty: iid \cite{GuSongEtAl2024a} (reported) & 0.5592  \\
+ Uncertainty: iid \cite{GuSongEtAl2024a} (reproduced)  & 0.6121  \\
{+ Uncertainty: LRPD (ours)}  & \textbf{0.6345}  \\ 
\midrule
\textit{Backbone: MapTRv2-CL \cite{MapTRv2}}  \\
+ Deterministic & 0.4789 \\
+ Uncertainty: iid \cite{GuSongEtAl2024a} (reported) & 0.4655  \\
+ Uncertainty: iid \cite{GuSongEtAl2024a} (reproduced) & 0.5204 \\
{+ Uncertainty: LRPD (ours)} & \textbf{0.5915} \\ 
\bottomrule
\end{tabular}%
\end{table}

\subsubsection{Trajectory Prediction Performance}
Table \ref{tab:traj_results} details the impact of map uncertainty on motion forecasting. Our evaluation reveals several key insights regarding the relationship between geometric uncertainty and prediction robustness. 

The bottom row of Table \ref{tab:traj_results}, representing the performance using ground-truth HD map data, serves as the empirical lower bound for prediction error. Notably, our proposed method with the MapTRv2-CL backbone achieves performance remarkably close to this lower bound, \textbf{only 1.9\%, 1.8\%, and 3.5\%} difference for the three metrics. 

\begin{table}[t]
\centering
\caption{Trajectory prediction performance for different backbones types and uncertainty estimation methods.}
\label{tab:traj_results}
\resizebox{\linewidth}{!}{%
\begin{tabular}{lccc}
\toprule
\textbf{Online Map Method} & \textbf{minADE}$_6$ $\downarrow$ & \textbf{minFDE}$_6$ $\downarrow$ & \textbf{MR}$_6$ $\downarrow$ \\ 
\midrule
\multicolumn{4}{l}{\textit{Backbone: MapTR \cite{liao2023maptr}}} \\
Deterministic & 0.4015 & 0.8418 & 0.0981 \\
+ Uncertainty: iid \cite{GuSongEtAl2024a} (reported) & 0.3854 & 0.7909 & 0.0834 \\
+ Uncertainty: iid \cite{GuSongEtAl2024a} (reproduced) & 0.3870 & 0.8036 & 0.0898 \\
+ Uncertainty: iid \cite{GuSongEtAl2024a} (reproduced) + score concat & 0.3875 & 0.7966 & 0.0852 \\
+ Uncertainty: iid \cite{GuSongEtAl2024a} (reproduced) + FiLM & 0.3771 & 0.7679 & 0.0797 \\
+ Uncertainty: iid \& Dual Dec. \cite{zhang2025delving}  & 0.3672 & 0.7395 & 0.0756 \\
+ BEV \cite{GuSongEtAl2024_BEV} & 0.3617 & 0.7401 & 0.0720 \\
+ BEV \& Dual Dec. \cite{zhang2025delving} & \textbf{0.3498} & \textbf{0.7021} & \textbf{0.0651} \\
{+ Uncertainty: LRPD (ours)} & {0.3573} & {0.7103} & {0.0652} \\
\midrule
\multicolumn{4}{l}{\textit{Backbone: MapTRv2 \cite{MapTRv2}}} \\
Deterministic & 0.4057 & 0.8499 & 0.0992 \\
+ Uncertainty: iid \cite{GuSongEtAl2024a} (reported) & 0.3930 & 0.8127 & 0.0857 \\
+ Uncertainty: iid \cite{GuSongEtAl2024a} (reproduced) & 0.3901 & 0.8103 & 0.0909 \\
+ Uncertainty: iid \& Dual Dec. \cite{zhang2025delving}  & 0.3670 & 0.7538 & 0.0708 \\
+ BEV \cite{GuSongEtAl2024_BEV} & 0.3844 & 0.7848 & 0.0741 \\
+ BEV \& Dual Dec. \cite{zhang2025delving} & \textbf{0.3475} & \textbf{0.6929} & {0.0661} \\
{+ Uncertainty: LRPD (ours)} & {0.3573} & {0.7167} & \textbf{0.0627} \\
\midrule
\multicolumn{4}{l}{\textit{Backbone: MapTRv2-CL \cite{MapTRv2}}} \\
Deterministic & 0.3790 & 0.7822 & 0.0853 \\
+ Uncertainty: iid \cite{GuSongEtAl2024a} (reported)  & 0.3727 & 0.7492 & 0.0726 \\
+ Uncertainty: iid \cite{GuSongEtAl2024a} (reproduced)  & 0.3659 & 0.7385 & 0.0764 \\
+ Uncertainty: iid \& Dual Dec. \cite{zhang2025delving}  & 0.3659 & 0.7404 & 0.0721 \\
+ BEV \cite{GuSongEtAl2024_BEV} & 0.3652 & 0.7323 & 0.0710 \\
+ BEV \& Dual Dec. \cite{zhang2025delving} & {0.3496} & {0.7096} & {0.0679} \\
{+ Uncertainty: LRPD (ours)} & \textbf{0.3423} & \textbf{0.6648} & \textbf{0.0555} \\
\midrule
{GT Map} & \textbf{0.3357} & \textbf{0.6525} & \textbf{0.0536} \\
\bottomrule
\end{tabular}%
}
\end{table}

As with the map generation experiments, we re-evaluated the code from \cite{GuSongEtAl2024a} and successfully reproduced the reported trends; however, our implementation achieved slightly lower prediction error in all cases except MapTR due to the optimized training schedule. This ensures that our LRPD gains are benchmarked against comparable and strongest possible version of the independent uncertainty baseline. The inclusion of uncertainty features consistently improves performance compared to deterministic baselines, as providing the prediction model with a measure of geometric reliability allows the network to better make use of predicted map elements. 

Furthermore, in the case of the MapTR backbone and diagonal uncertainty, we ablated the influence of the FiLM encoding of the confidence scores from Sec.~\ref{subsubsec: film} versus a simple concatenation of the scores (rows 4 and 5). Observing better performance in the case of FiLM, we used it with all the MapTR, MapTRv2 and MapTRv2-CL backbones. Thus, all the bottom rows in Tab.~\ref{tab:traj_results} use the procedure in Sec.~\ref{subsubsec: film}, in addition to the explicit encoding of the LRPD representation from Sec.~\ref{subsubsec: explicit_enc}.  

The results consistently demonstrate that structured uncertainty (LRPD) outperforms independent methods across all backbones. Specifically, on the MapTRv2-CL backbone, our method establishes a new state-of-the-art for online-map-based motion prediction with uncertainty.

Finally, we compare our approach against the "Dual Decoder" extension \cite{zhang2025delving} that utilizes significantly larger decoder heads. Despite having a more parameter-efficient head, our LRPD approach remains competitive or superior. Our structured-covariance formulation is inherently compatible with such architectural extensions, which represents a promising direction for future work. Overall, the highest prediction accuracy is achieved by combining our proposed LRPD formulation with the MapTRv2-CL backbone.


%% file: sections/05_conclusion.tex
\section{Conclusion}
\label{sec:conclusion}

We presented a structured probabilistic framework for online map generation using a Low-Rank plus Diagonal (LRPD) covariance parameterization. This approach captures spatial correlations and geometric constraints without the computational overhead of full covariance matrices.

Evaluations on nuScenes show that our formulation improves map generation metrics and establishes new state-of-the-art for map-based motion prediction. By closely approaching the performance of ground-truth HD maps, our results emphasize the importance of structured uncertainty in bridging the gap between online perception and high-definition priors.

Future research could expand the low-rank parameterization beyond intra-element dependencies to capture inter-element correlations, such as relationships between adjacent lanes. Additionally, integrating these structured uncertainty estimates downstream would enable safety-aware trajectory planning. Finally, as our LRPD formulation is architecture-agnostic, integrating it with advanced diffusion-based decoders \cite{sun2026diffsemanticfusion} remains an exciting avenue for future work.